\documentclass[10pt,twocolumn,letterpaper]{article}

\usepackage{cvpr}
\usepackage{times}
\usepackage{epsfig}
\usepackage{graphicx}
\usepackage{amsmath}
\usepackage{amssymb}
\usepackage{subfigure}
\usepackage{longtable}
\usepackage{multirow}
\usepackage{supertabular}
\usepackage{slashbox}
\usepackage[pagebackref=true,breaklinks=true,letterpaper=true,colorlinks,bookmarks=false]{hyperref}

\cvprfinalcopy % *** Uncomment this line for the final submission

\def\cvprPaperID{437} % *** Enter the CVPR Paper ID here

\ifcvprfinal\fi
\begin{document}

%%%%%%%%% TITLE
\title{EraseReLU: A Simple Way to Ease the Training of Deep Convolution Neural Networks}

\author{Xuanyi Dong$^1$, Guoliang Kang$^1$, Kun Zhan$^2$, Yi Yang$^1$\\
$^1$ CAI, University of Technology Sydney; $^2$ Lanzhou University
}

\maketitle
%\thispagestyle{empty}

%%%%%%%%% ABSTRACT
\begin{abstract}
For most state-of-the-art architectures, Rectified Linear Unit (ReLU) becomes a standard component accompanied with each layer. Although ReLU can ease the network training to an extent, the character of blocking negative values may suppress the propagation of useful information and leads to the difficulty of optimizing \textit{very deep} Convolutional Neural Networks (CNNs). Moreover, stacking layers with nonlinear activations is hard to approximate the intrinsic linear transformations between feature representations.

In this paper, we investigate the effect of erasing ReLUs of certain layers and apply it to various representative architectures following deterministic rules. It can ease the optimization and improve the generalization performance for \textit{very deep} CNN models. We find two key factors being essential to the performance improvement: 1) the location where ReLU should be erased inside the basic module; 2) the proportion of basic modules to erase ReLU; We show that erasing the last ReLU layer of all basic modules in a network usually yields improved performance. In experiments, our approach successfully improves the performance of various representative architectures, and we report the improved results on SVHN, CIFAR-10/100, and ImageNet. Moreover, we achieve competitive single-model performance on CIFAR-100 with 16.53\% error rate compared to state-of-the-art.
\end{abstract}

\section{Introduction}

Since the success of AlexNet~\cite{krizhevsky2012imagenet} in the ILSVRC-2012 competition~\cite{russakovsky2015imagenet}, more and more researchers move their focus on deep CNNs. Features learned from the neural networks significantly improve the performance of the large-scale vision recognition task, and can successfully be transferred to a large variety of computer vision tasks, such as object detection \cite{girshick2014rich}, pose estimation \cite{wei2016convolutional} and human-object interactions \cite{gkioxari2017interactnet}.
Due to the powerful transferability of CNN models, ``network engineering'' has attracted much research interest. This leads researchers to explore more effective and efficient network architectures.

\begin{figure}[t]
\center
\includegraphics[width=0.45\textwidth]{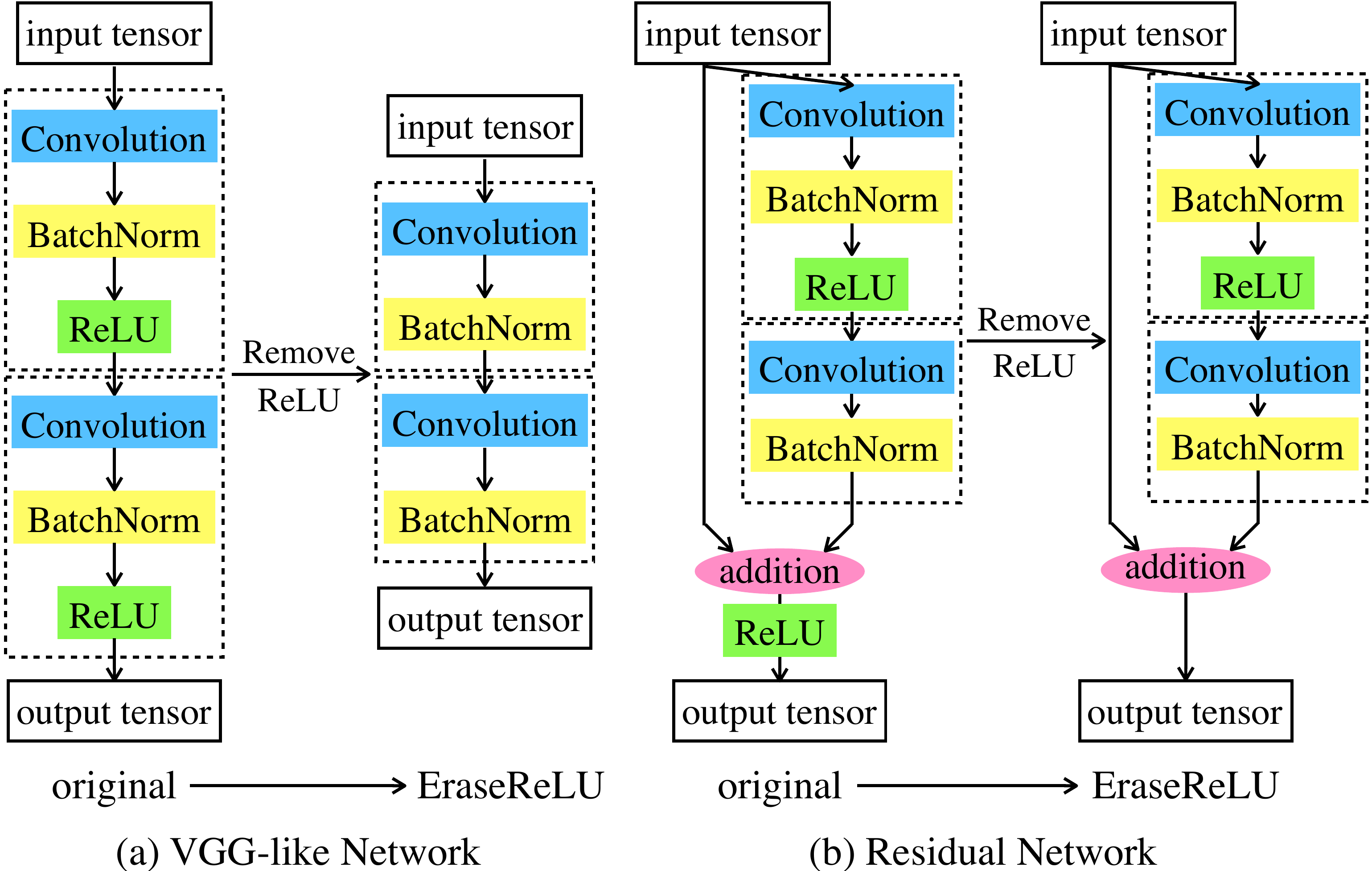}
\caption{(a) shows a part of VGG-like networks. (b) shows a block of ResNet. ``original'' and ``proposed'' mean the original architecture and the improved architecture by EraseReLU.
Our proposed model can achieve a higher performance compared to the original.
}
\label{fig:framework}
\end{figure}

The state-of-the-art CNN architectures become increasingly deeper and more complex. The VGGNet~\cite{simonyan2014very} extends the depth of AlexNet from eight to nineteen layers. The GoogleNet~\cite{szegedy2015going,szegedy2016rethinking} designs the Inception module explicitly by incorporating the multi-scale property into the architecture. ResNet~\cite{he2016deep} proposes a residual learning framework to ease the training of networks, and can successfully train networks with more than 1000 layers. DenseNet~\cite{huang2017densely} connects each layer to every other layer, which encourages feature reuse and substantially reduces the number of parameters. These different architectures share a key characteristic: they incorporate many nonlinear units, ReLU, in the networks.

ReLU layers are widely used in all CNN architectures. It has also been demonstrated to be more powerful compared to other nonlinear layers in most situations, such as sigmoid and tanh~\cite{glorot2011deep}. ReLU can not only increase the nonlinearity but also ameliorate gradient vanish/explosion phenomenon in CNNs. Therefore, a convolution layer or a fully-connected layer is usually accompanied with a ReLU layer by default in the state-of-the-art CNN architectures. \textit{Is this design principle necessary and helpful for the classification or some other vision tasks?}

We empirically find that reducing the nonlinearity in \textit{very deep} CNNs eases the difficulty of neural network training.
As networks go deeper, the benefits from depth and complexity become less~\cite{zagoruyko2017diracnets}.
For example, there is about 1.3\% accuracy improvement from ResNet-110 to ResNet-164 on CIFAR-10 \cite{krizhevsky2009learning}. However, six times deeper network, ResNet-1001, even decreases the accuracy about 1.7\%. In this way, the trained network is far from the capacity it should achieve. When we simply erase the last ReLU layer in residual blocks, the accuracy of a network with more than 1000 layers can still increase, whereas the \textit{very deep} ResNet tends to decrease the accuracy.

In this work, we propose a simple but effective method to improve the performance of deep CNN architectures by erasing the ReLU layers.
As shown in Figure~\ref{fig:framework}, our approach erases the last ReLU layer of the basic module in the deep neural network.
We thus call our approach ``EraseReLU''.
Our intuition is that some nonlinear layers may suppress information through the forward pass.
For example, in some cases~\cite{dong2017more}, the feature tensors after ReLU can be dominated by zero values, which harms the performance and cannot recover because the gradient of zero value is also zero.
Therefore, as our approach significantly reduces the nonlinearity in CNN models, it can help the information propagation in \textit{very deep} networks. EraseReLU also benefits the network optimization, which usually leads the model to converge faster in the early training epochs.
By investigating various factors of applying EraseReLU,
we find that two key factors to improve the performance.
1) the location where ReLU should be erased; 2) the proportion of modules to erase ReLU.
Moreover, we empirically demonstrate that erasing the last ReLU layer in all basic modules usually tends to result in performance improvement for various state-of-the-art CNN architectures, ResNet, Pre-act ResNet~\cite{he2016identity}, Wide ResNet~\cite{zagoruyko2016wide}, Inception-V2~\cite{ioffe2015batch} and ResNeXt~\cite{xie2017aggregated}.
Besides, we provide the theoretical analysis of EraseReLU which proves ReLU layers suppress the gradient back-prorogation in deep CNNs.

In summary, this paper makes the following contributions:
\begin{enumerate}
\itemsep0em 
\item We propose a simple but effective approach to improve the classification performance of \textit{very deep} CNN models by erasing the last ReLU layer of a certain proportion of basic modules in CNNs.
	  Moreover, the proposed approach can ease the difficulty of deep CNN training, which makes the CNN model converge faster.
\item We provide the theoretical analysis of our EraseReLU. We demonstrate gradients in \textit{very deep} CNN models are suppressed by ReLU layers.
\item We empirically show significant improvements on various of state-of-the-art CNN architectures. We use four benchmark datasets, including the large-scale dataset, ImageNet.
\end{enumerate}

\section{Related Work}

\textbf{Nonlinearity.}
The nonlinear unit plays an essential role in strengthening the representation ability of a deep neural network. In early years, sigmoid or tanh are standard recipes for building shallow neural networks. Since the rise of deep learning, ReLU \cite{glorot2011deep} has been found to be more powerful in easing the training of deep architectures and contributed a lot to the success of many record-holders \cite{krizhevsky2012imagenet,simonyan2014very,szegedy2016rethinking,szegedy2017inception,he2016deep}. There exist lots of variants of ReLU nowadays, such as Leaky ReLU \cite{maas2013rectifier}, PReLU \cite{he2015delving}, \emph{etc}. The common ground shared by these units is that the computation will be linear on a subset of neurons.  Models trained with such kinds of nonlinear units can be viewed as a combination of an exponential number of linear models which share parameters \cite{nair2010rectified}. This inspires us that modeling the local linearity explicitly may be useful. In this paper, we extend this linearity from the \textit{subset} to the \textit{full set} of neurons in some layer and empirically found that this extension effectively improves the performance of the model.

\textbf{Architecture.}
Krizhevsky won the ILSVRC-2012 competition and revealed that a large and deep CNN~\cite{krizhevsky2012imagenet} is capable of achieving benchmark results on a highly challenging dataset.
\cite{zeiler2014visualizing} proposes a novel visualization technique providing insight into the CNN features as well as a new architecture ZFNet.
NIN~\cite{lin2013network} leverages one-by-one convolutional layers to make the network become deeper and yield better performance.
VGGNet~\cite{simonyan2014very} further promotes the depth of CNN to 19 weighted layers resulting in a significant improvement.

Inception module is first proposed in~\cite{szegedy2015going}, which considers the Hebbian and multi-scale principle in CNN.
\cite{ioffe2015batch}~proposes the Batch Normalization (BN) to accelerate the network training. They also applied BN to a new variant of the GoogleNet, named as BN-Inception.
\cite{szegedy2016rethinking} proposes several general design principles to improve Inception module which leads a new CNN architecture, Inception-v3.
\cite{szegedy2017inception} combines the advantages of Inception architectures with residual connections to speedup the CNN training.
%\cite{kontschieder2015deep} trains classification trees with the representation learned from deep convolutional networks in an end-to-end manner.

Highway network~\cite{srivastava2015training} is designed to ease gradient-based training of very deep networks.
\cite{he2016deep} proposes the deep residual network that achieves a remarkable breakthrough in ImageNet classification and won the 1st places various of ImageNet and COCO competitions. The proposed residual learning can make the network easier to optimize and gain accuracy from considerably increased depth.
Following ResNet, \cite{he2016identity} proposes the Pre-activation ResNet improving the ResNet by using pre-activation block.
Wide ResNet~\cite{zagoruyko2016wide} decreases the depth and increases the width of residual networks. It tackles the problem of diminishing feature reuse for training very deep residual networks.
ResNeXt~\cite{xie2017aggregated} optimizes the convolution layer in ResNet by aggregating a set of transformations with the same topology.

Some researchers incorporate the stochastic procedure in the CNN models.
\cite{huang2016deep} proposes stochastic depth, a training procedure enabling the seemingly contradictory setup to train short networks and use deep networks at test time.
\cite{gastaldi2017shake} proposes to use parallel branches with a stochastic affine combination in ResNet to the avoid overfitting problem.
Our approach is a different way to improve CNN models compared with them, and we can complement each other.
While a contemporary work~\cite{zhao2017training} argues that 1:1 convolution and ReLU ratio is not the best choice to design the network architectures, we observe that only tuning the convolution and ReLU ratio may not always lead to improvement for different network structures.
Instead, the location where ReLUs should be erased is the key factor, which is the focus of this paper.

\section{Methodology}

The mapping function between different representations can be linear or non-linear. If the intrinsic relationship between two different representations is a linear mapping, it is hard to learn to approximate such kind of linear mapping through stacking of non-linear transformations, especially when the architecture is quite deep and the data is scarce.
For example, \cite{he2016deep} illustrates that optimization can be difficult in approximating an identity mapping by stacking multiple layers with non-linear activations in very deep neural networks.
ReLU layers introduce linearity for the subset of neurons with positive responses.
However, we find that for a subset of layers, it can be helpful to keep the linearity for the neurons with negative responses, \ie, explicitly forcing the linear mapping for a subset of layers.

%\cite{he2016deep} suggests that optimization can be difficult in approximating a linear transformation, identity mapping, by multiple nonlinear layers in very deep neural networks. If one hypothesizes that, in some layers of deep neural networks, linear transformations are optimum, then it may decrease performance by using nonlinear transformation. This motivates us to erase certain nonlinear units (ReLU) in deep neural networks.

\subsection{EraseReLU}

\begin{figure}[t]
\center
\includegraphics[width=0.45\textwidth]{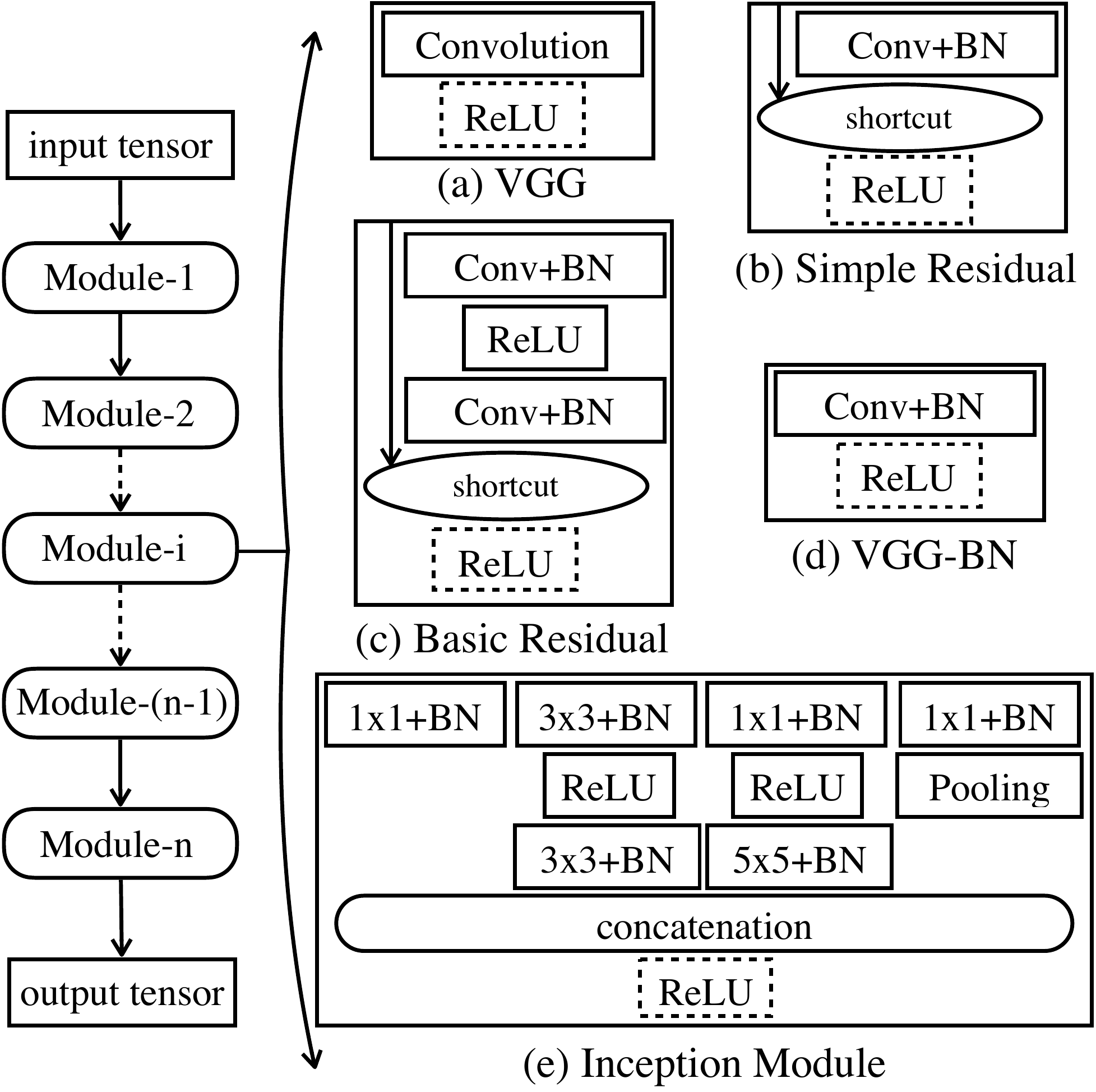}
\caption{In most CNN architectures, the feature extraction part is stacked by many similar modules with different configurations. (a) and (d) are the modules for VGG style network. (b) and (c) show the residual style modules. (e) shows an example of the Inception module. Our EraseReLU erases the ReLU with dashed boxes.
}
\label{fig:module}
\end{figure}

In most architectures, the network consists of multiple stacked core modules. Therefore, a network can be formulated~\footnote{For simplification, we ignore the fully-connected and pooling layers.} as :

\vspace{-3mm}
{\small
\begin{align}
F(x) = f_{n} \circ f_{n-1} \circ ... \circ f_{i} \circ ... \circ f_{2} \circ f_{1} \circ x ,
\end{align}
}
\vspace{-3mm}

\noindent where $f_{i}$ indicates the $i$-th basic unit in the network and $f_{i} \circ x$ equals $f_{i}(x)$. 
Figure~\ref{fig:module} illustrates five different kinds of $f$ in different typical CNN architectures.
As we stack more such modules in CNN, the network tends to overfit the training set and the optimization becomes more difficult.
Residual connection~\cite{he2016deep} can alleviate this phenomenon..
However, these two problems are still unsolved~\cite{zagoruyko2017diracnets}
We empirically find that reducing nonlinearities of $f$ can be helpful for ameliorating the overfitting and optimization problems in \textit{very deep} neural networks.

What is the most efficient way to reduce the nonlinearities of $f$ and maintain the model capacity at the same time?
There are usually three operations in CNN models: convolution, batch normalization, and ReLU.
Convolution operation is a linear unit and also essential for the model capacity, we thus do not change the convolution operation.
BN is not a linear unit strictly but can be approximately regarded as a linear unit. It can avoid the gradient explosion by stabilizing the distribution and reducing the internal covariate, we thus should also retain this operation. Therefore, there left two directions to reduce the nonlinearities, modifying ReLU or optimizing the module structure. The module structure has thousands of combinations of different operations, which is beyond the scope of our paper. Hereto, we eliminate all choices except modifying ReLU.

We can observe that modules in Figure~\ref{fig:module} share a common character, \ie, the last layer is a ReLU layer:

\vspace{-3mm}
{\small
\begin{align}
F(x) = Module(x) = ReLU( Module' (x) ),
\end{align}
}
\vspace{-3mm}

\noindent where most architectures also have this character. If we erase the last ReLU layer, we can preserve the overall structure.
On the contrary, if we erase the middle ReLU layer of these modules, it will destroy the module structure and decrease the module capacity, thus be harmful to the performance (See discussion in Sec.\ref{sec:connection}).
We empirically observe erasing the last ReLU layer in each module is capable of easing the training difficulty and considered as a regularization to improve the final performance.

\begin{figure}[!t]
\center
\subfigure[VGG-style Nets]{
\label{subfig:plain_ratio}
\includegraphics[width=0.224\textwidth]{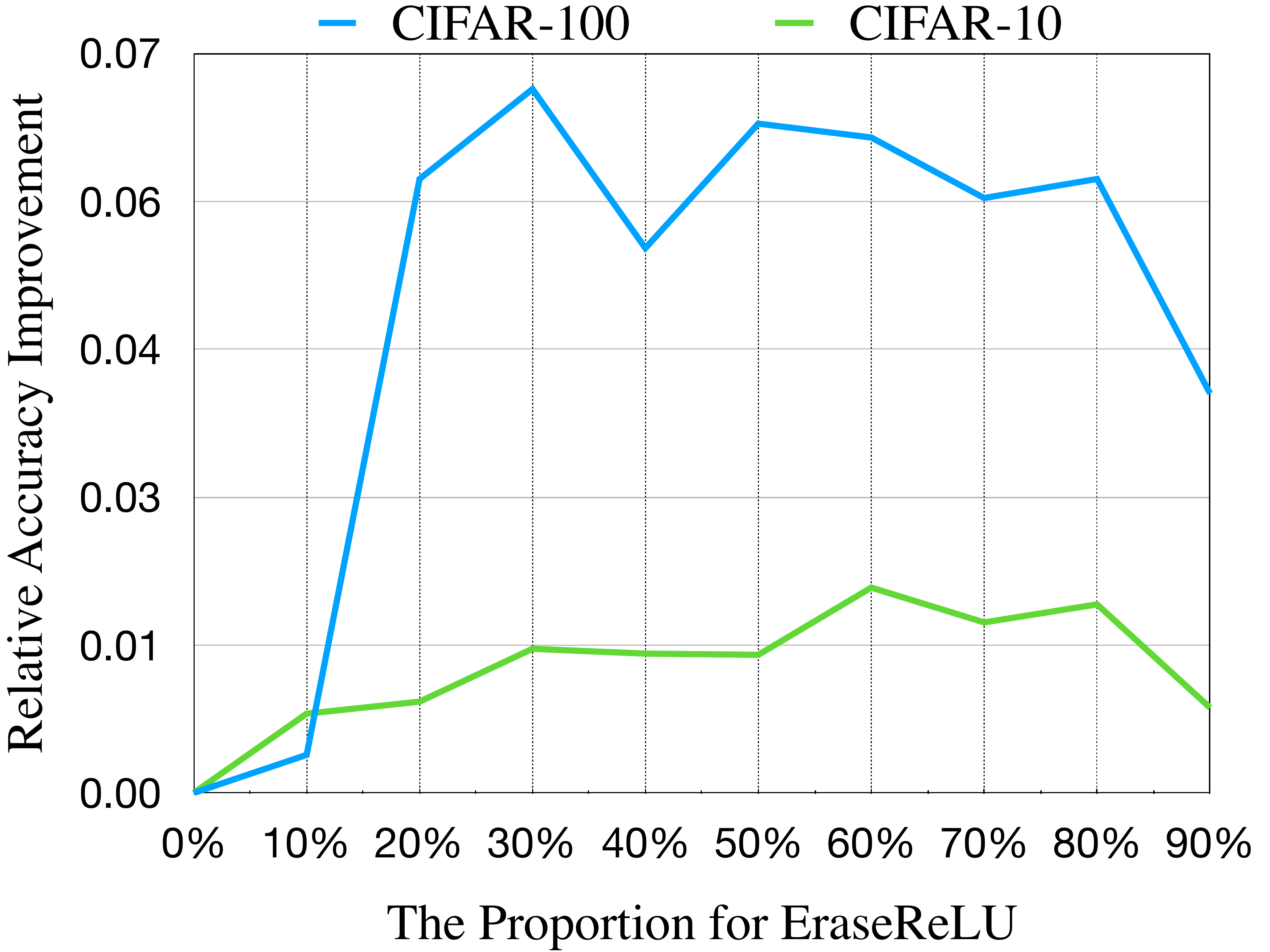}
}
\subfigure[Residual-style Nets]{
\label{subfig:res_ratio}
\includegraphics[width=0.224\textwidth]{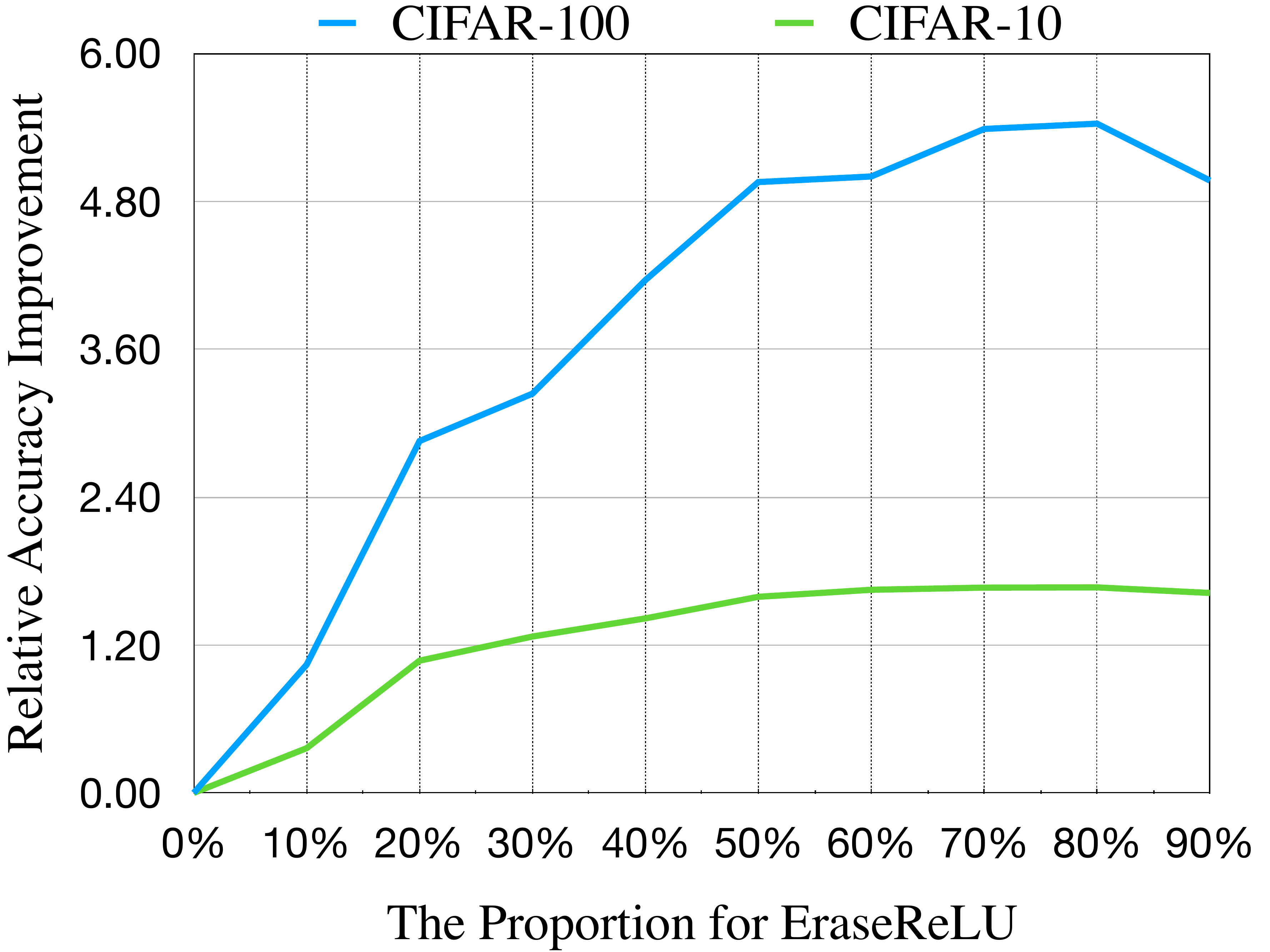}
}
\subfigure[Multiple Layer Perception Nets]{
\label{subfig:mnist}
\includegraphics[width=0.224\textwidth]{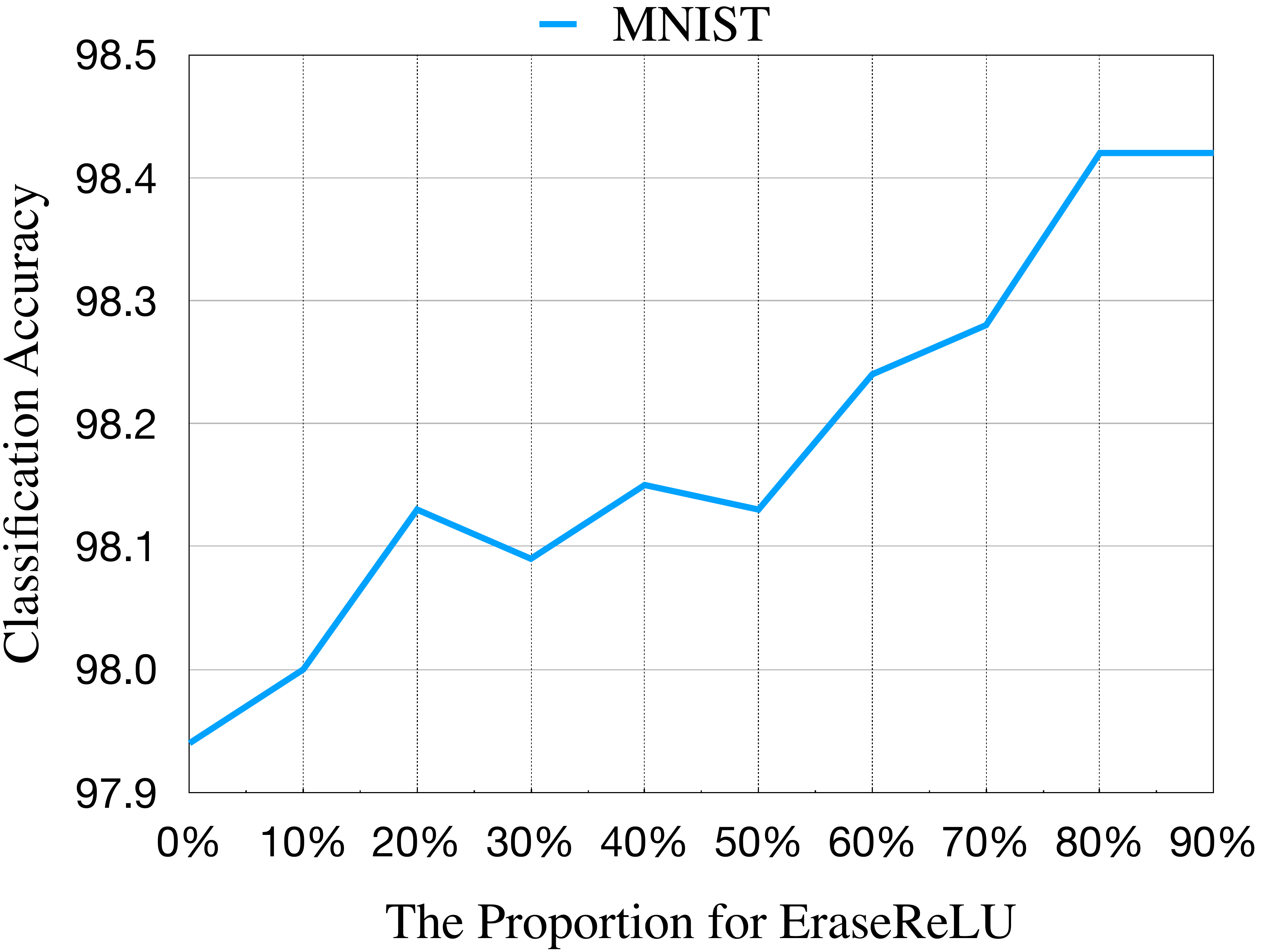}
}
\subfigure[VGG-style Nets]{
\label{subfig:depth}
\includegraphics[width=0.224\textwidth]{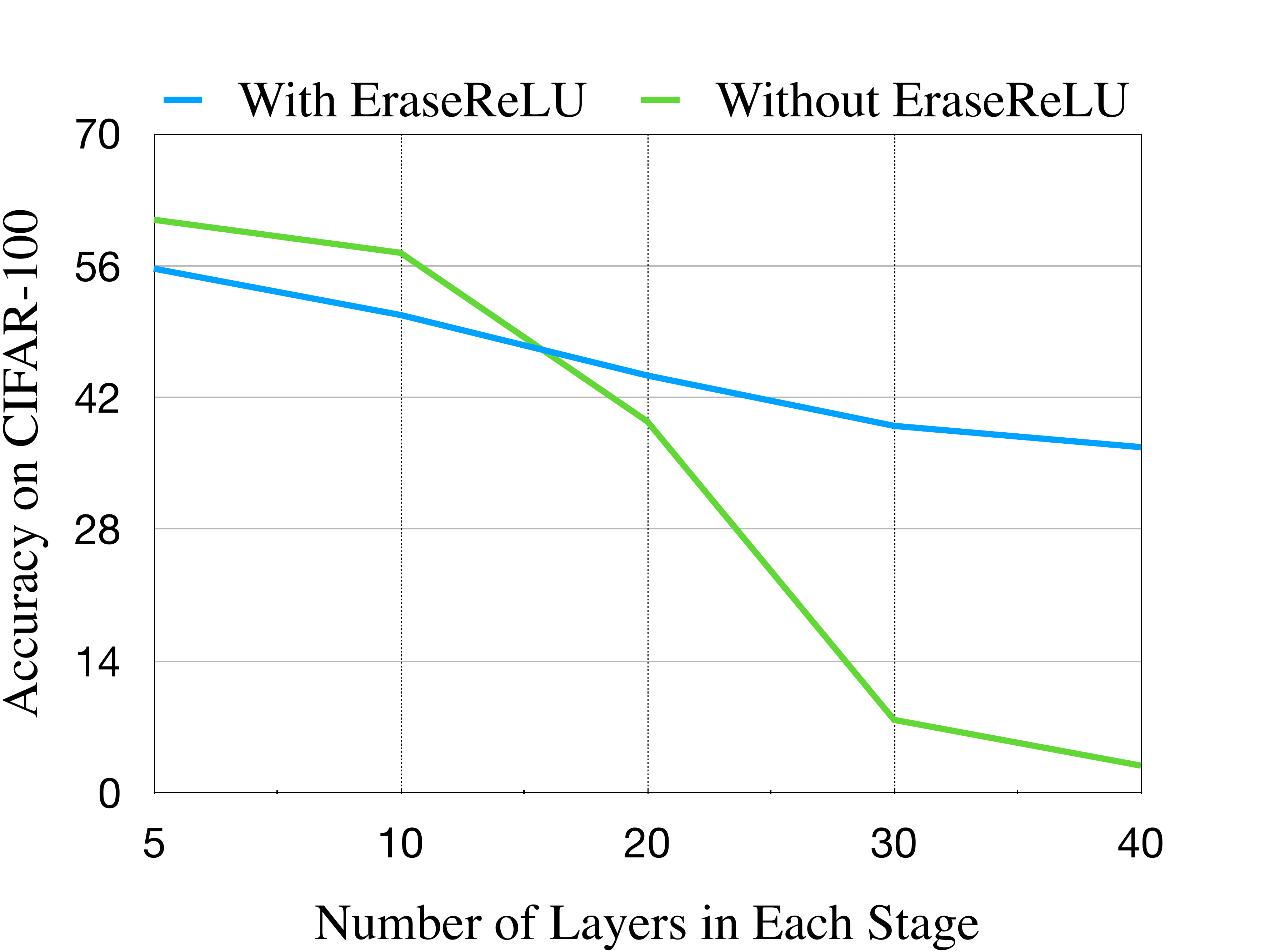}
}
\caption{Comparison of classification accuracy with and without EraseReLU.
(a) shows the relatively accuracy improvement of a VGG-style network with 31 weighted layers regarding the different proportion of modules to apply our EraseReLU.
(b) is similar as (a), whereas we replace the VGG-style network by a residual-style network.
In these two figures, the green and blue lines represent the results of CIFAR-100 and CIFAR-10, respectively.
(c) illustrates the absolute accuracy comparison of a multiple layer perception network with 12 weighted layers on MNIST, regarding the different proportion for EraseReLU.
(d) shows the accuracy on CIFAR-100 regarding different number of weighted layers, where the architecture is the same as in (a).
}
\label{fig:relu_ratio}
\end{figure}

Modules in Figure~\ref{fig:module} can be categorized as the after-activation structure~\cite{he2016identity}, where activation operations (BN \& ReLU) are after the convolutional layer.
Pre-activation structure~\cite{he2016identity,huang2017densely,zagoruyko2016wide} is another kind of module.
It moves BN+ReLU to the head of the convolution layer, and ReLU thus is not the tail of module.
To apply EraseReLU to these architectures, we first transfer them into the after-activation structure and then apply EraseReLU,
because the middle ReLU layer is essential for performance as we discuss before.

Hereto, the locations where ReLU layers should be erased have been discussed.
It is still not clear that EraseReLU should be applied to which module.
If we arbitrarily choose the module to apply EraseReLU, there exist thousands of choice combinations and some of them are even equivalent.
Therefore, we use the proportion of modules which should be applied with EraseReLU for efficiency.
Given a specific proportion, we uniformly sample the locations where the module to be applied with EraseReLU.
For example, if we apply EraseReLU on ResNet with the proportion of 50\%, then we will erase the last ReLU of the $1$-th, $3$-th, $5$-th and \etc modules in ResNet.
The location where ReLU should be erased and the proportion of modules to apply EraseReLU are two key factors for performance improvement.

\subsection{Analysis of the EreaseReLU's Effect}\label{sec:effect}

\begin{figure*}[t]
\center
\includegraphics[width=0.95\textwidth]{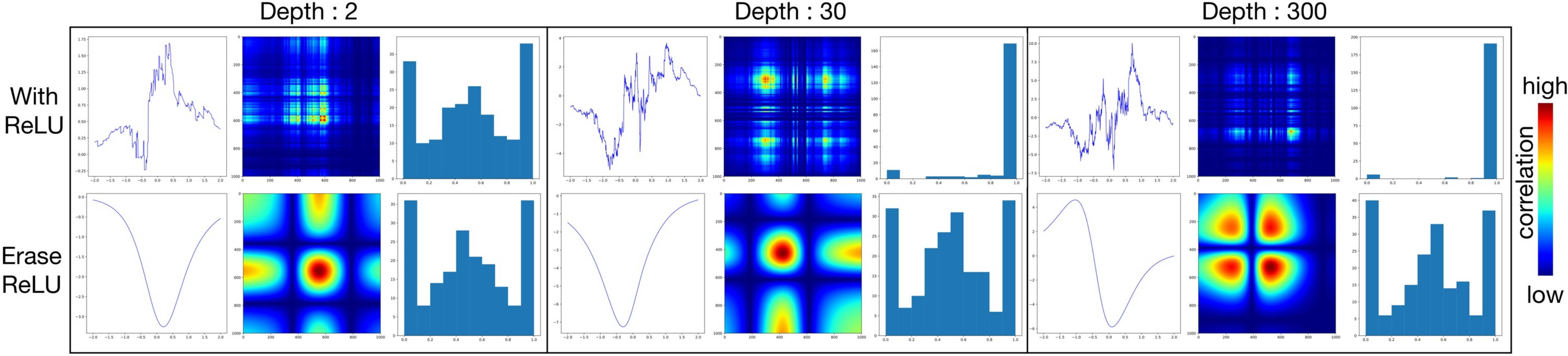}
\caption{Comparison between networks with and without EarseReLU regarding different depths.
In each triplet, the first figure show the gradients for $\text{input} \in [-2,2]$; the second figure show the visualization of gradients' covariance;
the third figure shows the distribution of average activation levels for each neuron in the last layer.
}
\label{fig:covariance}
\end{figure*}

In this section, we analyze the effect of different factors when applying EraseReLU to improve CNN architectures.
We perform two models stacked by thirty similar modules on CIFAR datasets.
One model uses the VGG-style module, and the other uses the residual-style module as shown in Figure~\ref{fig:module}.
Both of them use a 3-by-3 convolutional layer with 16 filters at the first layer. There are three stages following the first convolutional layer, where each stage has ten modules with the different number of output channels, \ie, 16, 32 and 64, respectively. Therefore, each of these two models has thirty-one weighted layers. Moreover, we try a multiple layer perception (MLP) network with 12 fully-connected layers. Each fully-connected layer in this MLP has one thousand output neurons followed by BN+ReLU+Dropout, except for the last one which maps the one thousand input neurons to ten neurons for MNIST classification. (More details can be found in supplementary materials)

%Figure~\ref{fig:module} shows relative accuracy improvement for the VGG-style and residual-based network.
%It can be found that a model with the proportion of 90\% for EraseReLU outperforms the original model (the proportion of 0\%) for a thirty-one weighted network.
Figure~\ref{subfig:plain_ratio} shows that reducing ReLU layers in a deep VGG-style network leads to the better performance.
It achieves the highest accuracy when the proportion for EraseReLU is 20\%.
Figure~\ref{subfig:res_ratio} performs the similar experiments as in Figure~\ref{subfig:plain_ratio}, while we replace the VGG-style network as the residual-style network, which dramatically eases the training procedure.
However, the original model is still inferior to the model with EraseReLU.
For this residual-style network, one with even a very small proportion of modules to erase ReLU (5\%) can outperform the original model.
Figure~\ref{subfig:mnist} demonstrates the same phenomenon in MLP, reducing the number of ReLU layers yields better performance.
Figure~\ref{subfig:depth} illustrates that the network trained with EraseReLU will gradually become better than the model without EraseReLU when the network goes deeper.
The module in the networks used in Figure~\ref{fig:relu_ratio} only has one ReLU layer.
If we use the proportion of 100\% for EraseReLU, there will be no nonlinearity in these models, which significantly reduce the performance, and we thus do not list the proportion of 100\% for EraseReLU in Figure~\ref{fig:relu_ratio}.

In practice, we usually use the proportion of 100\% for EraseReLU, which means that we apply EraseReLU on all modules.
There will usually have more than one ReLU layers in the module of state-of-the-art architectures.
Therefore, even the proportion of 100\% can maintain enough nonlinearity of these models.

\subsection{Theoretical Analysis}
Inspired by~\cite{balduzzi2017shattered}, we design a neural network $f_{w} : {\mathbb R} \rightarrow {\mathbb R}$ taking scalars to scalars.
It contains of $N$ modules, where the first layer and the last layer are fully-connected layers.
The middle module is similar to the simple residual module in Figure~\ref{fig:module}, whereas we replace the convolutional and BN layer with a fully-connected layer and a LayerNorm layer~\cite{ba2016layer}. Each hidden layer contains two hundred rectifier neurons.
At initialization, the function $f_{w}$ takes the input $x \in [-2, 2]$ in a 1-dim grid of 1000 data points. We initialize the weight and bias by \cite{he2015delving}.
Figure~\ref{fig:covariance} illustrates the gradient of the network for $x \in [-2, 2]$ and its covariance matrix in the first two subfigures of each triplet.
As the depth increases, the model using EraseReLU does not hurt the correlations between gradients.
On the contrary, the original model (denoted as ``With ReLU'') shows a decreased tendency of correlations between gradients.
The third sub-figure in each triplet of Figure~\ref{fig:covariance} shows the activation distribution of the last hidden layer.
If the neuron is greater than 0, it is activated; otherwise, it is deactivated.
For the model with ReLU, the distribution becomes increasingly bimodal with depth.
It decreases efficiency with which rectifier nonlinearities are utilized~\cite{balduzzi2017shattered}.
But the model with EraseReLU can maintain neurons to be utilized more efficiently even the depth becomes 300.

\begin{table*}[!t]
\centering
\setlength{\tabcolsep}{2.6pt}
\begin{tabular}{|c|c|c|c|c|c|c||c|c|c|} \hline
          & Depth&ResNet          & PReLU$^{all}$  & PReLU$^{sum}$  & ER*           & ER             & ER$\dagger$   & SDR             &{\bf ER$\dagger$+SDR} \\
       	  \hline
          & 20   &12.46 $\pm$0.32 & 11.18$\pm$0.17 & 12.16$\pm$0.18 &21.34$\pm$0.34 & 12.47$\pm$0.21 & 5.38$\pm$0.20 & 22.51$\pm$0.30 &24.09$\pm$0.09    \\
  %& 32   &5.45 $\pm$ 0.14 & 3.63 $\pm$0.09 & 5.17 $\pm$0.23 &16.29$\pm$0.12 & 4.29 $\pm$0.23 & 0.80$\pm$0.02 & 14.71$\pm$0.39 &15.76$\pm$0.14    \\
 Train    & 56   &2.38 $\pm$ 0.22 & 1.70 $\pm$0.24 & 0.89 $\pm$0.13 &13.28$\pm$0.79 & 0.86 $\pm$0.08 & 0.09$\pm$0.02 & 4.91 $\pm$0.12 & 4.95$\pm$0.03    \\
 -ing     & 110  &0.34 $\pm$ 0.05 & 0.21 $\pm$0.04 & 0.33 $\pm$0.01 &9.76 $\pm$0.04 & 0.16 $\pm$0.10 & 0.29$\pm$0.01 & 0.56 $\pm$0.02 & 0.50$\pm$0.04    \\
          \hline
          & 20   &32.85 $\pm$0.48& 33.40 $\pm$0.59 & 32.95$\pm$0.30 &35.15$\pm$0.36 & 32.29$\pm$0.33 &{\bf 31.82$\pm$0.30}& 32.87$\pm$0.15&33.24$\pm$0.36 \\
 %  & 32   &31.00 $\pm$0.08& 31.51 $\pm$0.44 & 31.86$\pm$0.07 &34.00$\pm$0.20 & 30.02$\pm$0.18 & 29.61$\pm$0.18 &{\bf 28.91$\pm$0.12}&29.44$\pm$0.22 \\
 Test     & 56   &30.86 $\pm$0.81& 30.60 $\pm$0.84 & 29.94$\pm$0.73 &33.64$\pm$0.33 & 28.56$\pm$0.17 & 27.23$\pm$0.01 & 25.60$\pm$0.36 &{\bf 25.01$\pm$0.22} \\
 -ing     & 110  &28.21 $\pm$0.46& 28.20 $\pm$0.98 & 27.45$\pm$0.46 &33.55$\pm$0.44 & 26.05$\pm$0.44 & 25.01$\pm$0.09 & 24.01$\pm$0.14 &{\bf 22.89$\pm$0.17} \\
          \hline
\end{tabular}
\vspace{2mm}
\caption{Classification error (\%) on CIFAR-100 with different variant of residual networks. ResNet represents the original residual network \cite{he2016deep}.
PReLU$^{all}$ indicates that we replace all ReLU layers by PReLU in ResNet. PReLU$^{sum}$ indicates that we only replace ReLU layers, which are right after the shortcut addition.
ER indicates that we apply EraseReLU on the last ReLU layer in the residual block.
ER* indicates the location where ReLU is erased is the first ReLU layer in the residual block.
SDR represents the stochastic depth ResNet~\cite{huang2016deep}.
As SDR train the model with 500 epoch, ER$\dagger$ is ER with the same training settings as SDR.
SDR can also be complementary to our approach, and we show the results of the ER trained by SDR as ER$\dagger$+SDR.
We run each model five times and show ``mean $\pm$ std''.
}
\vspace{-2mm}
\label{table:depth}
\end{table*}

\subsection{Connection with other non-linear activations}\label{sec:connection}
There exist many variants of ReLU, \eg, Leaky ReLU and PReLU. They aim to alleviate the gradient vanishing problem. 
While these variants may benefit the training of deep neural networks compared to ReLU, the performance improvement is negligible~\cite{xu2015empirical}.
% Inspired by ReLU that inducing linearity for a subset of neurons, we explicitly inducing the linear mapping 
% \textit{i.e.} erasing ReLUs, for a subset of layers. Experiments illustrate that EraseReLU outperforms ReLU with a large margin.
EraseReLU can be considered as a variant of ReLU. It extends identity mapping from positive domain to the whole domain, but leads to a significant improvement compared to other ReLU variants.
Table~\ref{table:depth} compares our EraseReLU with other variants of ReLU and stochastic depth ResNet~\cite{huang2016deep} (denoted as SDR).
We can find that PReLU achieves a much lower error on the training data, yet obtains a higher testing error compared with EraseReLU.
Given enough training data, PReLU can theoretically learn an identity mapping if our EraseReLU is an optimal solution.
But in practical, EraseReLU always yield a much better performance than PReLU.

ER* in Table~\ref{table:depth} indicates erasing the middle ReLU layer in residual blocks.
The middle ReLU layer is essential for as well as the final accuracy.
ER* destroys the overall structure of the basic module, and it thus significantly reduces the performance compared to the original model.
Instead, ER, which erases the last ReLU layer, yields the consistent performance improvement.
SDR is a technique to boost the CNN training when the network is deep, which can complement our EraseReLU.
By combining the mutual benefits from SDR and EraseReLU, we achieve 22.89 error rate on ResNet-110, which improves the SDR by about relative 5\% and ResNet by about 19\%.

\section{Experiments}

% In this section, we investigate the performance of the proposed method.
% We empirically demonstrate the effectiveness of our proposed approach on various state-of-the-art architectures.
% These typical CNN models with EraseReLU are evaluated on four datasets, SVHN~\cite{netzer2011reading}, CIFAR-10, CIFAR-100~\cite{krizhevsky2009learning} and ImageNet-1k~\cite{russakovsky2015imagenet}.

\subsection{Datasets}

\textbf{SVHN.} 
The Street View House Numbers (SVHN) dataset is obtained from house numbers in Google Street View images. This dataset contains ten classes, from digit 0 to digit 9.
There are 73257 color digit images in the training set, as well as the 26032 images in the testing set. It also has 531131 additional, somewhat less difficult samples, to use as extra training data.
Following the common experiment setting on SVHN, such as~\cite{zagoruyko2016wide,huang2017densely}, we only use the official training and testing data. When training models, we only divide the pixel values by 255 to range into [0,1] without any data augmentation.

\textbf{CIFAR.}
The CIFAR-10 dataset consists of 60000 images categorized into ten classes.
There are 50000 training images with 5000 images per class, and 10000 testing images with 1000 images per class.
The CIFAR-100 dataset is similar as CIFAR-10 but contains 100 classes. We use the official training and testing sets of these two datasets.
Following the common practice \cite{he2016deep,xie2017aggregated,huang2017densely}, we normalize the data using the means and standard deviations of RGB channels. We also adopt the random horizontal flip and random crop\footnote{Pad 4 pixels on each border and randomly crop a 32x32 region} data argumentations.

\textbf{ImageNet.}
The ILSVRC classification dataset contains 1000 classes. There are about 1.2 million images for training, and 50000 for validation. We use the same data argumentation as in \cite{xie2017aggregated,huang2017densely} for training. For evaluation, we only use the single-crop with input image size of $224^{2}$.
We also generate three subsets of ImageNet for empirical studies. These three subsets randomly sample 10\%, 20\% and 30\% images from the ImageNet dataset. Therefore, all of them have 1000 classes for training and testing images, but the number of images is different. We refer these three datasets as ImageNet-10\%, ImageNet-20\% and ImageNet-30\%. To be noticed, ImageNet-10\% is a subset of ImageNet-20\%, and ImageNet-20\% is a subset of ImageNet-30\%.

\subsection{Experiments on SVHN and CIFAR}

In this section, we apply our approach to improve five different architectures, ResNet, Pre-act ResNet, Wide ResNet, ResNeXt and Inception-V2.
We demonstrate the comparison on SVHN, CIFAR-10, and CIFAR-100.

{\bf Experiment settings.}
The proportion for EraseReLU is a hyper-parameter, which is selected from \{25\%, 50\%, 75\%, 100\%\} according to validation sets.
As the Inception-V2 was designed for the ImageNet with the input image size of $224^{2}$, we design a special Inception-V2 model for CIFAR datasets. It is shown as modules in Figure~\ref{fig:module}, where the $5\times5$ convolutional layer is replaced by two $3\times3$ convolutional layers.
The proportions of output channels from the $1\times1$, $3\times$, $5\times5$ convolution branches and the pooling branch are $1:8:2:1$ in one Inception module.
Our Inception-V2 on CIFAR has a convolutional layer with 64 output channels at first, followed by three Inception stages. Each stage contains ten Inception modules, where the base output channels are 16, 32 and 64, respectively.

\begin{figure}[t]
\center
\subfigure[Accuracy (L) and Loss (R) of ResNet-164 on CIFAR-10]{
\label{subfig:resnet110}
\includegraphics[width=0.47\textwidth]{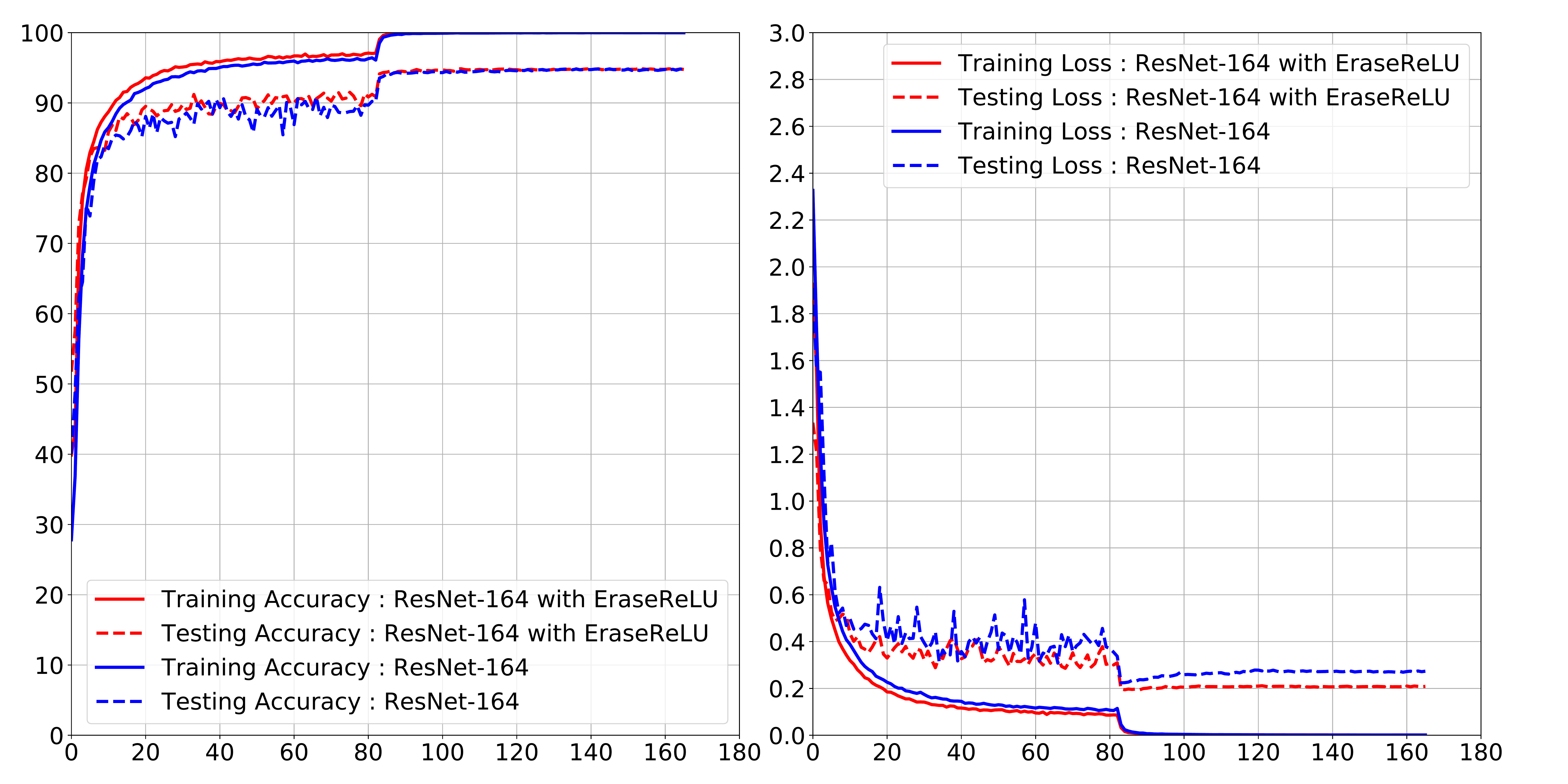}
}
\subfigure[Accuracy (L) and Loss (R) of ResNet-164 on CIFAR-100]{
\label{subfig:resnet164}
\includegraphics[width=0.47\textwidth]{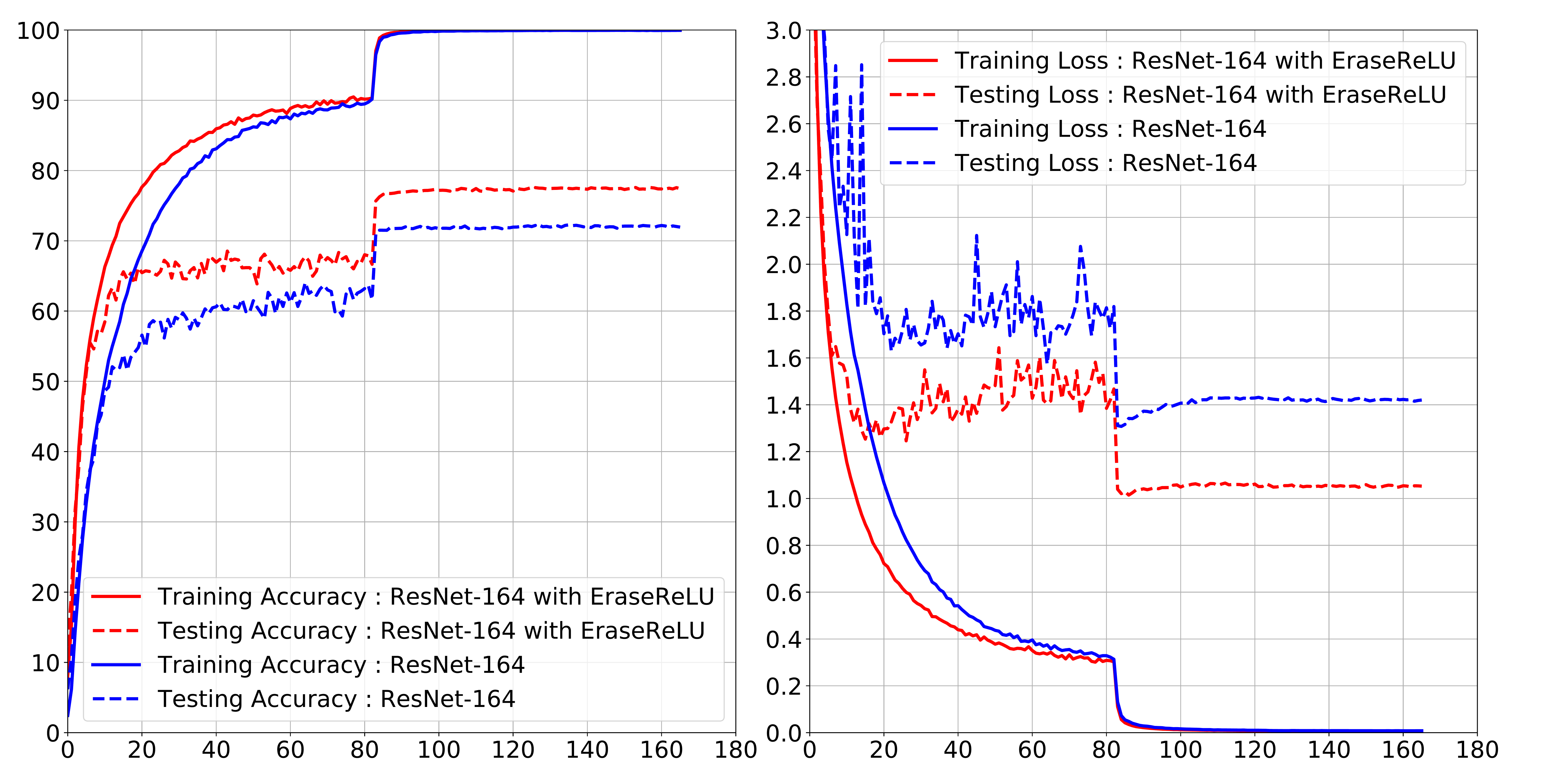}
}
\caption{Comparison between EraseReLU and the original model. In each subfigure, the x-axis indicates the train epoch and the y-axis means the accuracy or loss. The red and blue lines are the model with EraseReLU and without EraseReLU.
The solid and dashed lines show the performance of the training and testing data.
}
\label{fig:resnet_cifar100}
\end{figure}

According to the validation, we use the proportion of 100\% for EraseReLU on ResNet, Pre-act ResNet, Wide ResNet and ResNeXt.
For Inception-V2, we use the proportion of 50\% for EraseReLU.
We the Inception-V2 model following the same training and testing strategies as in~\cite{zagoruyko2016wide}.
For all other networks, we follow the same training and testing strategies as in their official papers~\cite{he2016deep,he2016identity,zagoruyko2016wide,xie2017aggregated}.
We run each model five times and report the mean error.

\begin{table*}[t]
\setlength{\tabcolsep}{2pt}
\begin{center}
\begin{tabular}{|c|c|c|c|c|c|c|c|c|c|}
\hline
\multirow{2}{*}{Architecture}& \multirow{2}{*}{Setting} & \multicolumn{3}{c|}{CIFAR-10}             & \multicolumn{3}{c|}{CIFAR-100} & \multicolumn{2}{c|}{SVHN}     \\ 
                                &        & original     &\cite{zhao2017training}& \bf{EraseReLU} &  original & \cite{zhao2017training} &  \bf{EraseReLU} &  original & \bf{EraseReLU}\\ \hline
\multirow{3}{*}{ResNet\cite{he2016deep}}
								&56      &     6.97     &  -          &  6.23 $\pm$ 0.21 & 30.60$\dagger$&  -        &28.56 $\pm$0.17 &  -        &  -   \\
						        &110     &     6.43     &  -          &  5.78 $\pm$ 0.10 & 28.21$\dagger$&  -        &26.05 $\pm$0.44 &  -        &  -   \\
                                &1202    &     7.93     &  -          &  4.96            &       -       &  -        &    -           &  -        &  -   \\ \hline
\multirow{3}{*}{Pre-act ResNet\cite{he2016identity}}
								&62      &6.98$\dagger$ &6.03$\pm$0.23&6.07$\pm$0.06&  29.44$\dagger$ &27.81$\pm$0.18& 27.54 $\pm$ 0.04        &  -        &  -   \\
								&110     &     6.37     &  -          &  5.72 $\pm$ 0.04 &     27.20     &26.59$\pm$0.10&25.76 $\pm$0.07 &  -        &  -   \\
						        &164     &     5.46     &  -          &  4.65            &     24.33     &25.88      &   22.41       &  -        &  -   \\
                                &1001    &     4.92     &  -          &  4.10            &     22.71     & -         &    20.63       &  -        &  -   \\\hline
\multirow{8}{*}{Wide ResNet\cite{zagoruyko2016wide}}
								&40-4    &     4.97     &  -          &  4.27            &     22.89     &  -        &   20.07       &  -        &  -   \\
                                &28-10   &     4.00     &  -          &  3.78            &     19.25     &  -        &   19.10       &  -        &  -   \\
                                &28-10-D &     3.89     &  -          &  3.88            &     18.85     &  -        &   18.60       &  -        &  -   \\
                                &52-1    &     6.43     &  -          &  5.37            &     29.89     &  -        &   26.45       & 2.08      & 1.87 \\
                                &52-1-D  &     6.28     &  -          &  5.65            &     29.78     &  -        &   25.00       & 1.70      & 1.54 \\
                                &52-10   & 3.69$\dagger$&  -          &  3.69            &19.19$\dagger$ &  -        &   18.19       & 1.81      & 1.64 \\\hline
\multirow{2}{*}{ResNeXt\cite{xie2017aggregated}}
								&29-8X64 &     3.65     &  -          &  3.57            &     17.77     &  -        &   17.23       &  -        &  -   \\
						        &29-16X64&     3.58     &  -          &\bf{3.56}         &     17.31     & -     &\bf{\color{blue}16.53}&  -    &  -   \\\hline
Inception-V2                    & 30     & 5.45$\dagger$&  -          &  5.29            & 28.29$\dagger$&  -        &   24.34       &  -        &  -   \\\hline
\end{tabular}
\end{center}
\caption{Classification error (\%) on CIFAR and SVHN datasets. The ``original'' means the architecture described in the original papers, and the EraseReLU means the architecture by applying our EraseReLU method on the original one. $\dagger$ indicates results run by ourselves. The setting formats follow the original papers, except that the setting of Inception-V2 indicates there are 30 Inception modules in the network. By applying EraseReLU, most of the architectures achieve lower error rates while using the same parameters and computation cost. We run each model five times and report ``mean ($\pm$ std)''.
`-' indicates they do not report results.}
\label{table:cifar}
\end{table*}

{\bf Experimental Analysis.}
Figure~\ref{fig:resnet_cifar100} illustrates the accuracy and loss curves of the original ResNet-110/ResNet-164 and the models using EraseReLU.
We can find that erasing ReLU layers in such deep neural networks can not only achieve higher accuracies but also reduce the overfitting of the training data. When the learning rate reduces from 0.1 to 0.01, the training loss goes down, but the testing loss goes up. The phenomenon is caused by over-fitting. The loss of original ResNet-110 rises from 1.2 to 1.5, while the model using EraseReLU just rises from 1.1 to 1.3. For the ResNet-164 model, by using EraseReLU the loss just rises about 0.01, but the original model raises more than 0.1. Therefore, the model using EraseReLU results in a much smaller loss increase than the original model. Moreover, with the network going deeper, we can obtain more benefits from EraseReLU. For example, the improvement of EraseReLU on ResNet-164 is much more than ResNet-110.

We demonstrate comparisons on SVHN and CIFAR in Table~\ref{table:cifar}.
For ResNet, EraseReLU improves the accuracy compared to the original architectures. The ResNet with 1202 layers achieves a worse performance than the network with 110 layers, but ours can still maintain the accuracy improvement even when the depth becomes very deep.
For Pre-act ResNet, we use both after-activation and EraseReLU. For Pre-act ResNet with 1001 layers on CIFAR-100, the error drops from 22.71\% to 20.07\% by our EraseReLU.
We can observe more than 2\% accuracy gain on Pre-act ResNet with 1001 layers.
As shown in \cite{he2016identity}, the after-activation style does not improve the performance. Thus the main contribution of the accuracy improvement is from EraseReLU.
For ResNeXt, the last ReLU of all residual blocks are removed, and we obtain about 0.8\% improvement for ResNeXt-29-16x64 on CIFAR-100.
The previous state-of-the-art result on CIFAR-100 is DenseNet-BC-190-40, 17.18\% error rate. We achieve a better result by applying EraseReLU on a model, which performs worse than DenseNet-BC-190-40.
On SVHN, EraseReLU improves the Wide ResNet by about 10\% relative accuracy. On CIFAR-10, we can observe general improvements on various kinds of networks. The Wide ResNet-52-10 with EraseReLU does not improve the performance, because our model may be close to the lower bound for the CIFAR-10 dataset.
Compared to \cite{zhao2017training}, we achieve more superior performance on CIFAR, and they lack the exploration on other different kinds of CNN architectures rather than ResNet.
Moreover, we will show the results on the large-scale dataset in the following section, whereas they only experiment on small-scale datasets.

\subsection{Experiments on ImageNet}

%We evaluate the performance of EraseReLU in terms of different architectures on the ImageNet classification task, and compare with the original models.
In this section, we first experiment on the ImageNet subsets to analyze the effect of model complexity and data scale for our EraseReLU.
We then evaluate our approach on the full ImageNet set to demonstrate the effectiveness.
For all experiments in this section, we use the official training code provided by PyTorch~\footnote{\url{https://github.com/pytorch/examples/blob/master/imagenet/main.py}} and train each network with 120 epochs with the learning rate initialized by 0.1. The learning rate is divided by 10 every 30 epochs.

\begin{figure}[!t]
\center
\subfigure[ResNet-101]{
\label{subfig:resnet101_imagenetdata}
\includegraphics[width=0.47\columnwidth]{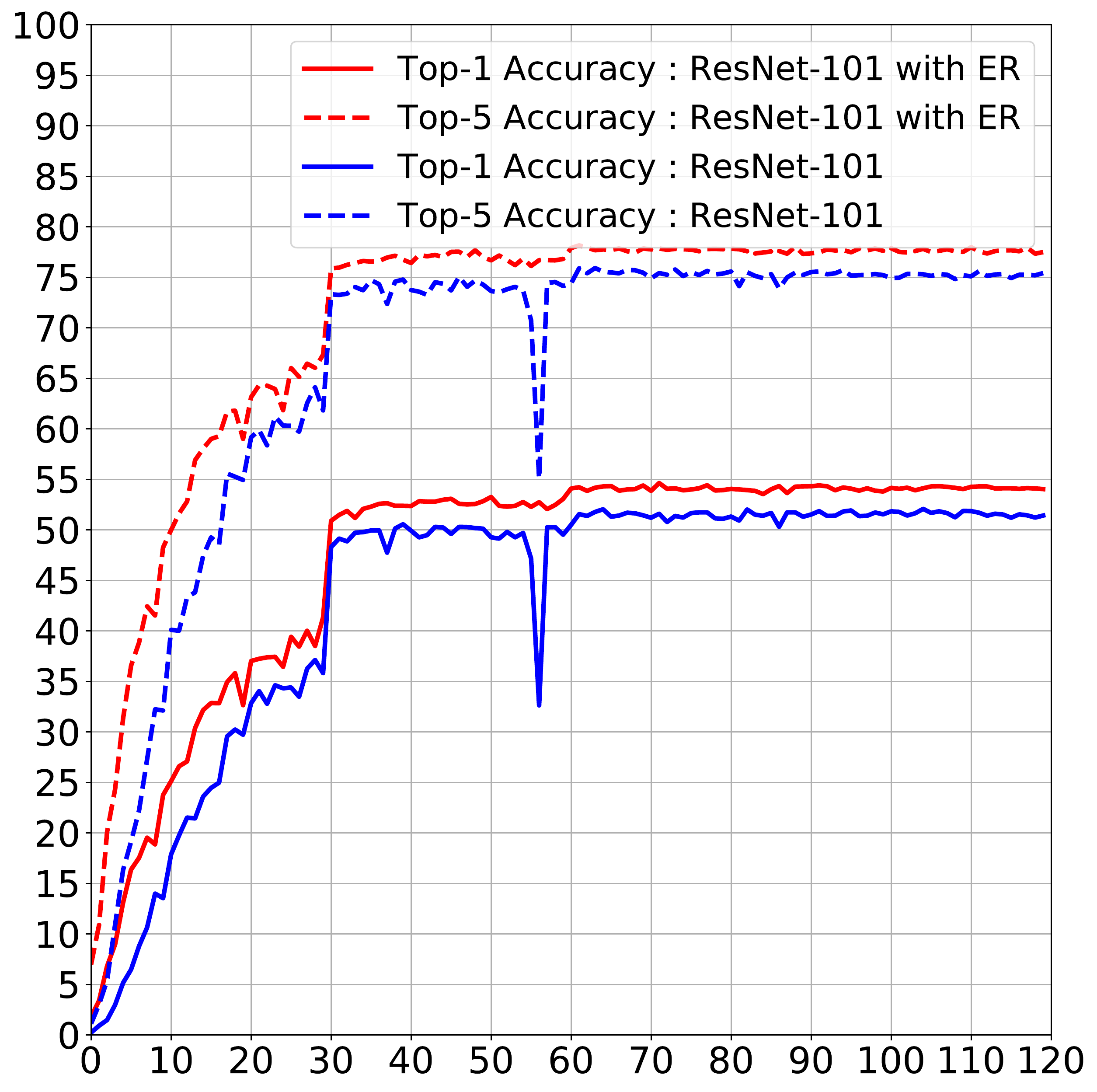}
}
\subfigure[ResNeXt-152 (32x4d)]{
\label{subfig:resnext152_imagenetdata}
\includegraphics[width=0.47\columnwidth]{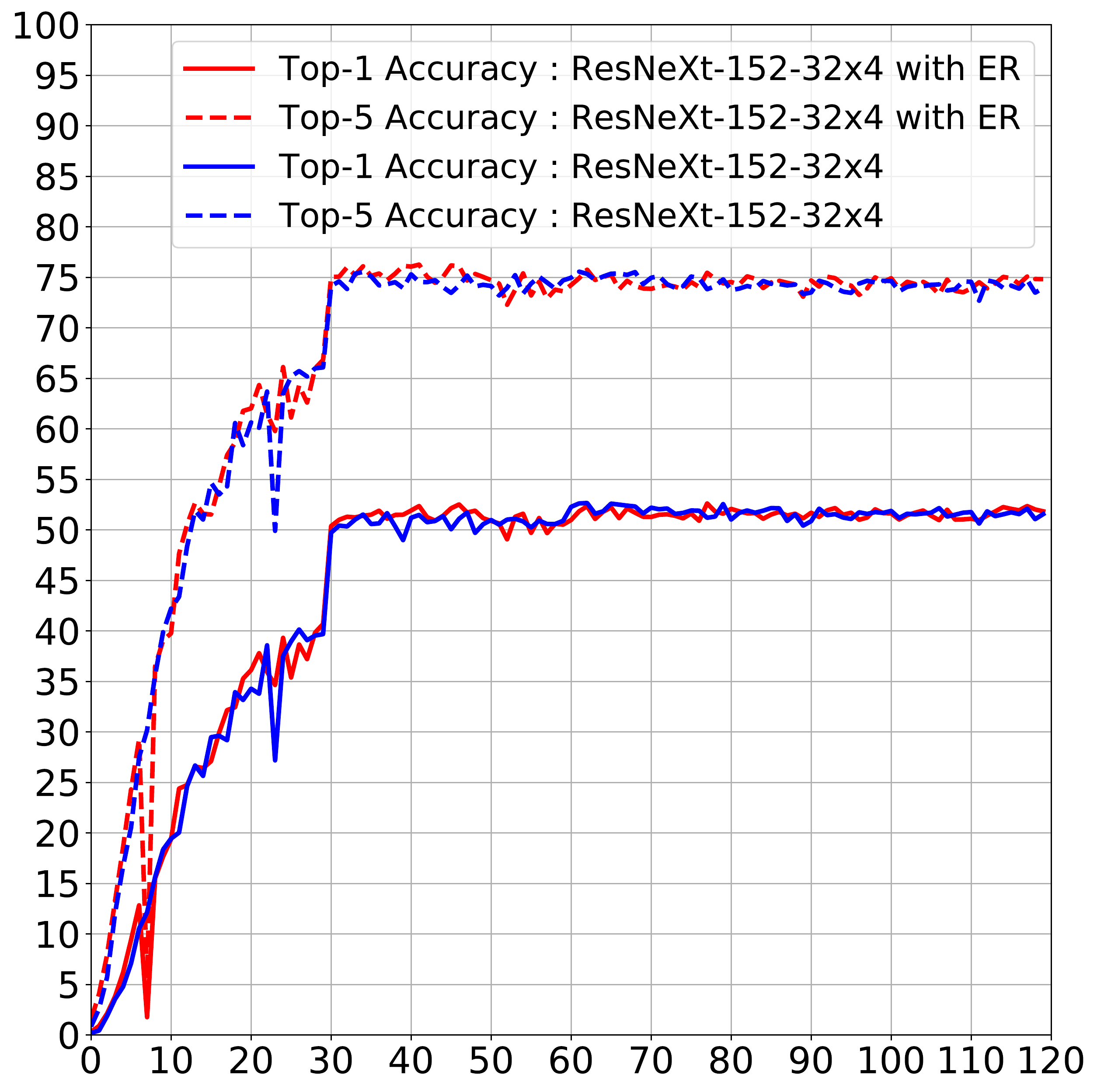}
}
\caption{Comparison of models with and without EraseReLU on the ImageNet-10\% in terms of the top-1 and top-5 accuracy (\%). The first subfigure shows the results of ResNet-101, and the second one shows ResNeXt-152 (32x4d). The blue and red lines indicate the original network and the network using EraseReLU, respectively. The solid and dashed lines show the top-1 and top-5 accuracy, respectively.
}
\label{fig:imagenet_subdata}
\end{figure}

Figure~\ref{subfig:resnet101_imagenetdata} compares ResNet-50 and ResNet-50 using EraseReLU with the proportion of 100\% on ImageNet-10\%.
Our EraseReLU outperforms the original model by about 3\% top-1 accuracy.
%On the subsets of ImageNet, we apply EraseReLU on ResNet-50, ResNet-101 and ResNeXt-152 (32x4).
%For ResNet-50 and ResNet-101, we simply erase the last ReLU activation layer for all residual block.
%For ResNeXt-152 (32x4), we only erase the last ReLU layer of every two neighboring residual blocks.
Figure~\ref{subfig:resnext152_imagenetdata} illustrates the results of ResNeXt-152 (32x4) using EraseReLU with the proportion of 50\%.
EraseReLU obtains a comparable  result compared to the original model.
%This phenomenon implies that EraseReLU may also work as a regularizer that benefits the model's generalization performance, especially when data is scarce.

Table~\ref{table:ratio} shows the comparison of the original ResNet-50 and the model using EraseReLU with the proportion of 100\%.
EraseReLU improves the performance of ResNet-50 on ImageNet-10\% about 1.4\% absolute top-1 accuracy.
When the numbers of training and validation images increase to ImageNet-20\%, the performance improvement goes down to 0.24\%.
On ImageNet-30\%, we also achieve a comparable results by using EraseReLU with the proportion of 25\%.
This indicates that the proportion is a key factor on large-scale dataset, which is not much sensitive on small-scale datasets.

On the ImageNet full set, we apply EraseReLU to two models, ResNet-152 and ResNet-200. To ensure a fair comparison, we adopt the public Torch implementation for ResNet \footnote{\url{https://github.com/facebook/fb.resnet.torch}} and only replace the model definition by the model applying our EraseReLU.
Therefore, all other factors are eliminated, such as data pre-process and data argumentations. We keep all the experiment settings the same as those used for ResNet.
For both ResNet-152 and ResNet-200, we use the EraseReLU with the proportion of 50\%.
EraseReLU improves the ResNet-152 on ImageNet dataset by 0.6\% top-1 accuracy.
For ResNet-200, we achieve 21.4\% top-1 accuracy outperforming the original one about 0.4\%. The ResNet-200 with EraseReLU also outperforms than the Pre-act ResNet-200, which is an improved version of ResNet.

We demonstrate EraseReLU can improve deep models in the large-scale dataset.
We do not apply EraseReLU to more sophisticated models on ImageNet because it needs much more hardware resources, which is unaffordable for us.
All the compared models in experiments are the state-of-the-art CNN architectures, which have been optimized by many researchers and validated by many systems/papers. It is not easy to remove their components.
If one removes a layer or a component arbitrarily, the performance will drop significantly.

\begin{table}[t]
\centering
\begin{tabular}{|c|c|c|c|c|} \hline
   \multicolumn{2}{|c|}{Sample Ratio}      &  10\%   &  20\%  &  30\%  \\\hline
\multirow{2}{*}{ResNet-50}         & top-1 & 47.60   & 36.76  & 31.11  \\
                                   & top-5 & 23.68   & 15.65  & 11.65  \\\hline
\multirow{2}{*}{ResNet-50 with ER} & top-1 & 46.18   & 36.42  & 31.10  \\
                                   & top-5 & 23.20   & 14.99  & 11.64  \\\hline
\end{tabular}
\vspace{2mm}
\caption{Top-1 and top-5 error rate on ImageNet validation subsets using different number of training images.
10\%, 20\% and 30\% indicates ImageNet-10\%, -20\% and -30\%, respectively.
For EraseReLU on ImageNet-10\% and -20\%, we use the proportion of 100\%.
For ImageNet-30\%, we use the proportion of 25\%.}
\vspace{-2mm}
\label{table:ratio}
\end{table}

\begin{table}[t]
\centering
\begin{tabular}{|c|c|c|} \hline
\backslashbox{Model}{Error}& top-1(\%)         & top-5(\%)        \\\hline
ResNet-152                 & 22.2              & 6.2              \\
ResNet-152 with ER         & 21.6              & 5.8              \\\hline
ResNet-200 $\dagger$       & 21.8              & 6.1              \\ 
Pre-act ResNet-200         & 21.7              & 5.8              \\
ResNet-200 with ER         & {\bf{21.4}}       & 5.8 \\\hline
\end{tabular}
\vspace{2mm}
\caption{Top-1 and top-5 error rate on ImageNet full validation set.
We compare various residual models with and without EraseReLU.
We use the proportion of 100\% for EraseReLU on ResNet-152 and Pre-act ResNet-200, referred as ``ER''.
They are evaluated on the ImageNet validation set.
We only use single-crop for testing with $224\times224$ image size. $\dagger$ indicates results run by ourselves.
}
\vspace{-2mm}
\label{table:imagenet}
\end{table}

\section{Conclusion}

In this paper, we investigate the effect of erasing ReLUs in deep CNN architectures following deterministic rules.
We find two key factors to performance improvement:
1) the location where the ReLU should be erased inside the basic module;
2) the proportion of modules to erase ReLU.
By leveraging these two factors, we propose a simple but effective approach to improve CNN models, named ``EraseReLU".
It can lead to a non-negligible improvement of classification performance because its effectiveness in easing the optimization and regularizing the training of deep neural networks.
Our approach improves the classification performance beyond various CNN architectures, such as ResNet, Pre-act ResNet, Wide ResNet and ResNeXt.
Most of them achieve a much higher performance while having the same computation cost compared to the original models.
We obtain more than 2\% absolute accuracy improvement on CIFAR-100 compared to the Pre-act ResNet-1001.
Our approach also leads 0.6\% accuracy improvement on the large-scale dataset ImageNet compared to ResNet-152.

{\small
\bibliographystyle{ieee}
\bibliography{egbib}
}

\end{document}